\documentclass[a4paper,fleqn]{cas-sc}

\usepackage[numbers]{natbib}
\usepackage{multirow}
\usepackage{graphicx}
\usepackage{amsmath}
\usepackage{amsxtra}
\usepackage{threeparttable}

\def\tsc#1{\csdef{#1}{\textsc{\lowercase{#1}}\xspace}}
\tsc{WGM}
\tsc{QE}
\tsc{EP}
\tsc{PMS}
\tsc{BEC}
\tsc{DE}

\begin{document}
\let\WriteBookmarks\relax
\def\floatpagepagefraction{1}
\def\textpagefraction{.001}
\shorttitle{An Efficient Hardware-Oriented Dropout Algorithm}
\shortauthors{YJ YEOH et~al.}

\title [mode = title]{An Efficient Hardware-Oriented Dropout Algorithm}                      



\author{Y.J. YEOH}
\cormark[1]
\ead{yeoh.yoeng-jye542@mail.kyutech.jp}


\address{Kyushu Institute of Technology, Japan}

\author{T. MORIE}
\ead{morie@brain.kyutech.ac.jp}

\author{H. TAMUKOH}
\ead{tamukoh@brain.kyutech.ac.jp}


\cortext[cor1]{Corresponding author}


\begin{abstract}
This paper proposes a hardware-oriented dropout algorithm, which is efficient for field programmable gate array (FPGA) implementation.
In deep neural networks (DNNs), overfitting occurs when networks are overtrained and adapt too well to training data.
Consequently, they fail in predicting unseen data used as test data.
Dropout is a common technique that is often applied in DNNs to overcome this problem.
In general, implementing such training algorithms of DNNs in embedded systems is difficult due to power and memory constraints.
Training DNNs is power-, time-, and memory- intensive; 
however, embedded systems require low power consumption and real-time processing.
An FPGA is suitable for embedded systems for its parallel processing characteristic and low operating power;
however, due to its limited memory and different architecture, it is difficult to apply general neural network algorithms.
Therefore, we propose a hardware-oriented dropout algorithm that can effectively utilize the characteristics of an FPGA with less memory required.
Software program verification demonstrates that the performance of the proposed method is identical to that of conventional dropout, and hardware synthesis demonstrates that it results in significant resource reduction.

\end{abstract}




\maketitle

\section{Introduction}
Deep neural networks (DNNs) have demonstrated promising performance in numerous applications, such as data mining, automation, and natural language processing \cite{intro, intro2, dnndata, dnnnlp}.
Examples include GoogLeNet-, which succeeded in image recognition in the ImageNet Large Scale Visual Recognition Challenge (ILSVRC) \cite{ilsvrc1, ilsvrc2}; 
and, Alpha-Go which defeated the top human player in the game Go \cite{alphago}. 
Advances in computer technology and networks have led to breakthrough in DNNs, which have become a popular topic among researchers. 
The number of studies on deep learning increases yearly.
As neural networks become increasingly deeper, the computational costs, memory and power consumption increase.
These factors make training more difficult, and it also become difficult to implement DNNs in an embedded system, which requires high mobility and real-time processing \cite{fpbnn}.

In the field of robotics and automation, mobility and power consumption have become quite important.
Researchers generally focus on the implementation of DNNs in hardware devices, which consume less power than software devices such as a CPU or graphics processing unit (GPU) \cite{fpbnn, reconfigurefpga}.
A field Programmable Gate Array (FPGA) is a digital hardware device composed mainly of transistors and programmable wires that enable flexible implementation and parallel processing \cite{fpga, fpga2}.
Unlike software devices that process serially, an FPGA enables parallel processing, making real-time processing a promising goal  \cite{fpga3}.
However, a large drawback of the FPGA is its memory constraint.
The memory resources in an FPGA are limited, and computations are executed digitally.
Thus to process a floating point number calculation or analog input from sensors, implementation becomes difficult.
Consequently, research has been conducted on modifying the algorithm of general DNNs, such as the binarized neural network (BNN) and XNOR net, to simplify the calculation, reduce the computational cost and memory resources, and make the algorithm suitable for hardware implementation \cite{bc, bnn, xnor, fpbnn}.
In addition to research on algorithms, there has also been research on architecture configuration design of FPGAs to achieve a parallel, resource-saving implementation \cite{reconfigurefpga}. 
This research includes designing, controlling, and manipulating the utilization of block RAM, digital signal processing (DSP), and look-up table (LUT) \cite{reconfigurefpga}.
Although many studies have focused on accelerating neural networks with an FPGA, most implemented only the inference phase, as the learning and update process of a DNN involves complex computation and high resource cost \cite{fpgainference, reconfigurefpga}.

\begin{figure}
\centering
\includegraphics[width=8cm]{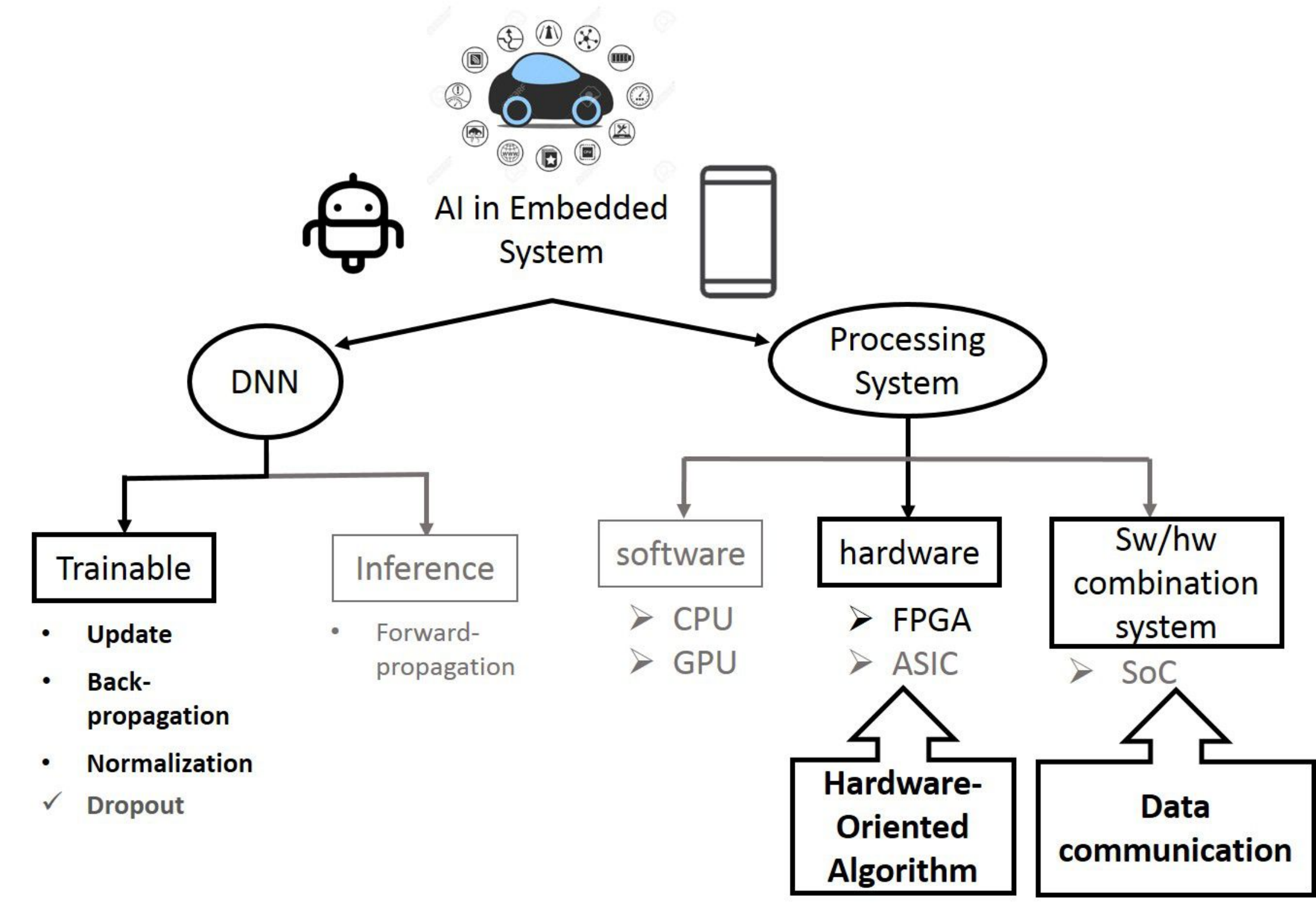}
\caption{Overview of research target.}
\label{f.overview}
\end{figure}

As illustrated in Fig. \ref{f.overview}, this paper focuses on developing a hardware-oriented algorithm for a trainable DNN in an FPGA, and we propose a hardware-oriented dropout algorithm for efficient implementation.
While training DNNs, the hyperparameters of the networks and the size of the dataset are difficult to initialize.
The larger and deeper the networks are, the more features must be trained and learned.
This may cause the networks to become too specified to the training data and learn even the noise.
As a result, overfitting occurs and resulting the networks fail during testing.
However, networks that are too shallow and small are also unable to train well, also resulting in poor performance.
As the size of a dataset decreases, the networks tend to adapt to all the data, thus overfitting to the data.
In contrast, when the dataset size is too large, the learning process become difficult and is too generalized.
The optimal hyperparameters and size of a dataset are difficult to determine, as they have no baseline and are case-dependent.

Dropout is a technique that is often applied to DNNs to solve the overfitting problem \cite{dropout}. 
By randomly dropping neurons with a certain ratio, it prevent the neurons from collaborating with each other \cite{dropout, dropout2}.
In addition, forcing part of the neuron to acquire a zero value simplifies the calculation.

However, to randomly drop neurons, random number generators (RNGs) are required, which are costly in terms of hardware.
Our proposed method eliminates the use of RNGs, substituting them with simple operators for performing dropout.
This results in saving hardware resources for other purposes.
The contributions of this paper are summarized as follows:
\begin{itemize}
	\item An efficient hardware-oriented algorithm for the dropout technique is proposed.
	\item Evaluation of the proposed method is performed with feedforward neural networks and recurrent neural network.
	\item A comparison between general dropout and the proposed method is performed.
	\item Hardware synthesis is performed for resource analysis.
\end{itemize}

\section{Related Works}
\subsection{Feedforward neural networks}
A feedforward neural network is generally used in image classification tasks.
This network is also the oldest and simplest neural network.
As implied by its name, the input is directly fed forward to the output, passing through a hidden layer with the sum of products of the weights and activation functions. 
In this network, there is no looping or feedback.
A multi-layer perceptron (MLP) and convolutional neural network (CNN) which are used in this study, are categorized as feedforward neural networks.

\subsubsection{Multi-layer perceptron (MLP)}

\begin{figure}
\centering
\includegraphics[width=8cm]{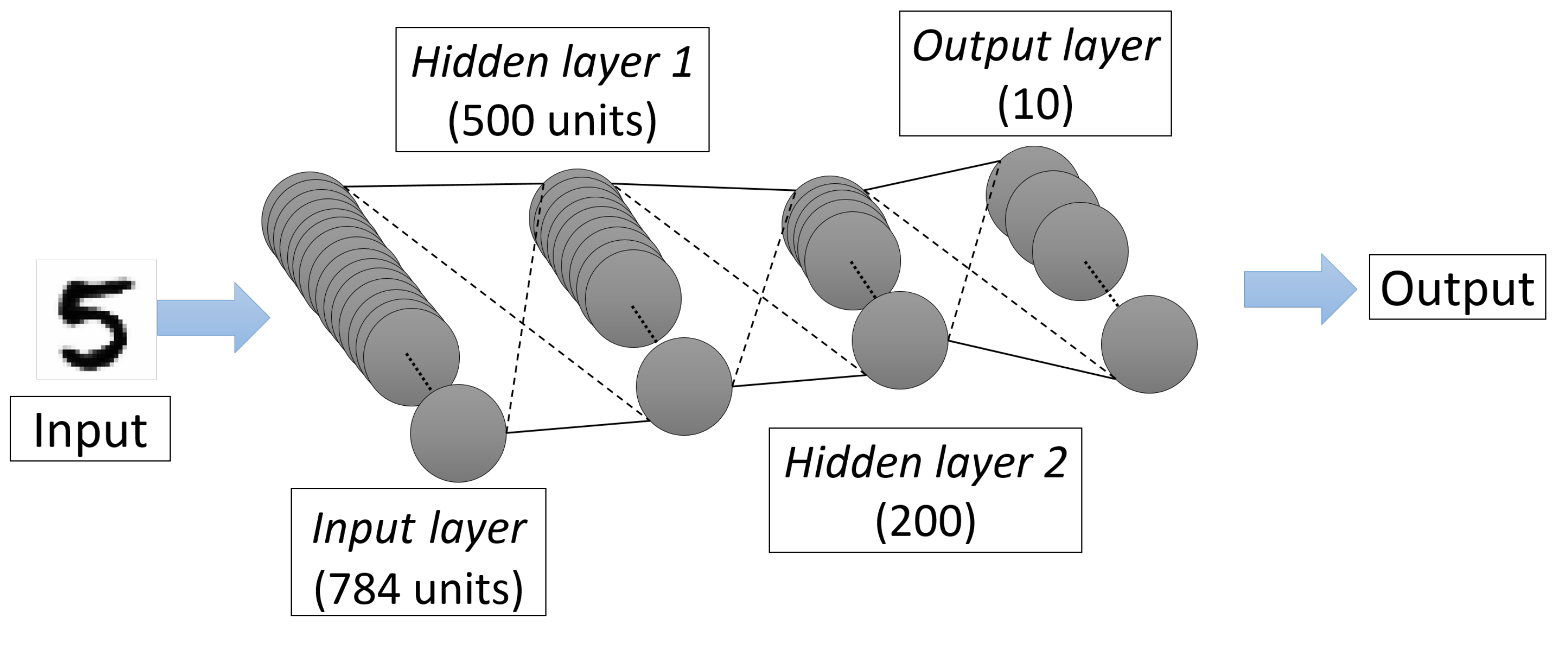}
\caption{Four-layers multi-layet perceptron with 500 and 200 hidden units.}
\label{f.mlp}
\end{figure}

An MLP is a simple neural network that maps the inputs to outputs with multiple hidden layers in between \cite{mlp}.
All neurons are fully connected by weights. 
The sum of the products of the weights $W$ and inputs $x$ is computed and fed to the next layer after the activation function, $f$.
Figure \ref{f.mlp} illustrates the four-layers MLP used in this work, with 500 and 200 hidden units in the two hidden layers respectively.

\subsubsection{Convolutional neural network (CNN)}

\begin{figure}
\centering
\includegraphics[width=8cm]{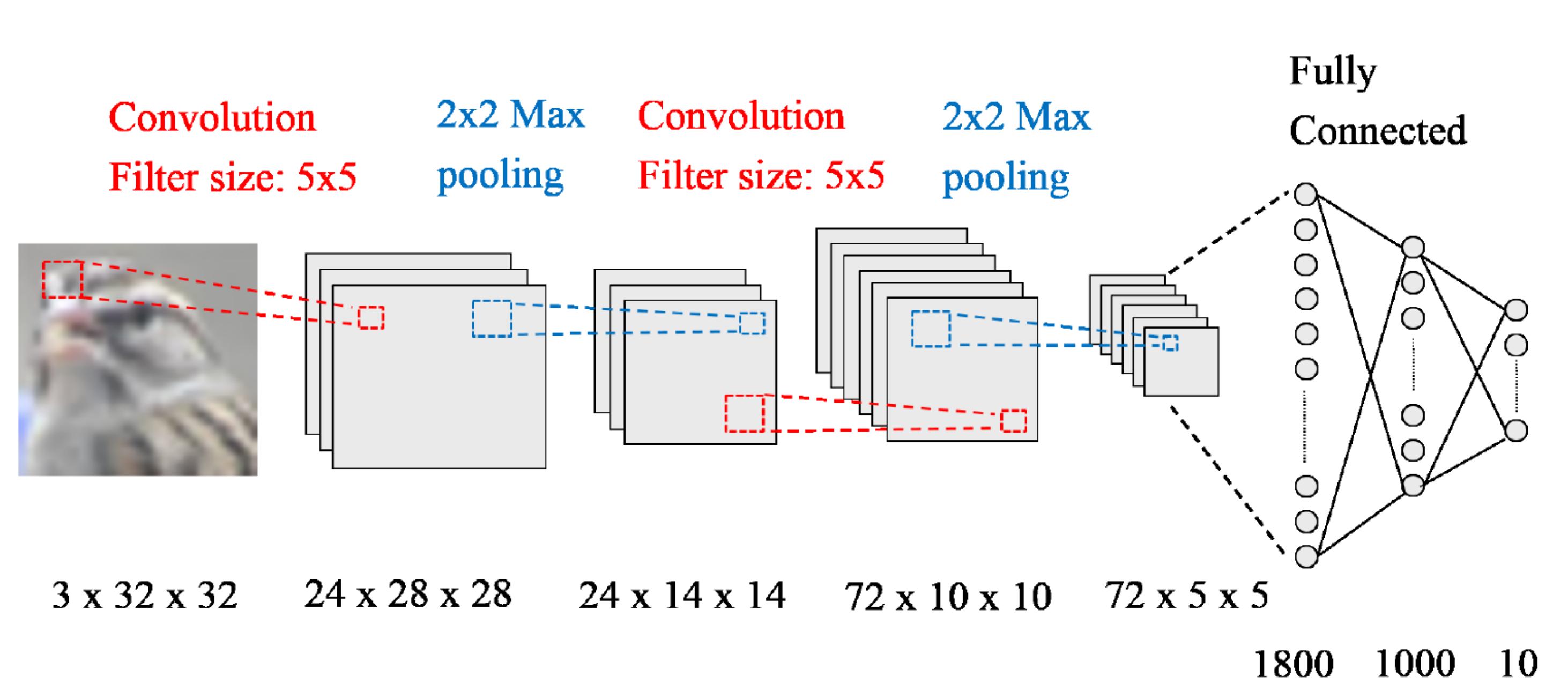}
\caption{Example of convolutional neural network: LeNet.}
\label{f.cnn}
\end{figure}

The first CNN was introduced in 1998 and has become popular in recent years due in part to its excellent performance in image recognition \cite{cnn}
In 2012, CNNs significantly reduced error rates and successfully outperformed other models in the ImageNet Large Scale Visual Recognition Challenge (ILSVRC) \cite{ilsvrc1}.
Deeper CNNs such as GoogLeNet, ResNet, and VGGNet were introduced in subsequent years.
In a CNN, input is computed by convolving with multiple filters, also known as kernels. 
The filters convolve part of the input and slide to the ascending part until all the input has passed through.
The weights are shared spatially.
The pooling layer in a CNN reduces the resolution of the input to a smaller scale, allowing the extracted features to have scale invariant properties.
CNNs usually consist of pairs of convolutional-pooling layers and fully-connected layers as a classifier, as illustrated in Fig. \ref{f.cnn}.
Figure \ref{f.cnn} presents LeNet which consists of two pairs of convolutional-pooling layers. 

\subsection{Recurrent neural network language model}

\begin{figure}
\begin{center}
\includegraphics[width=8cm]{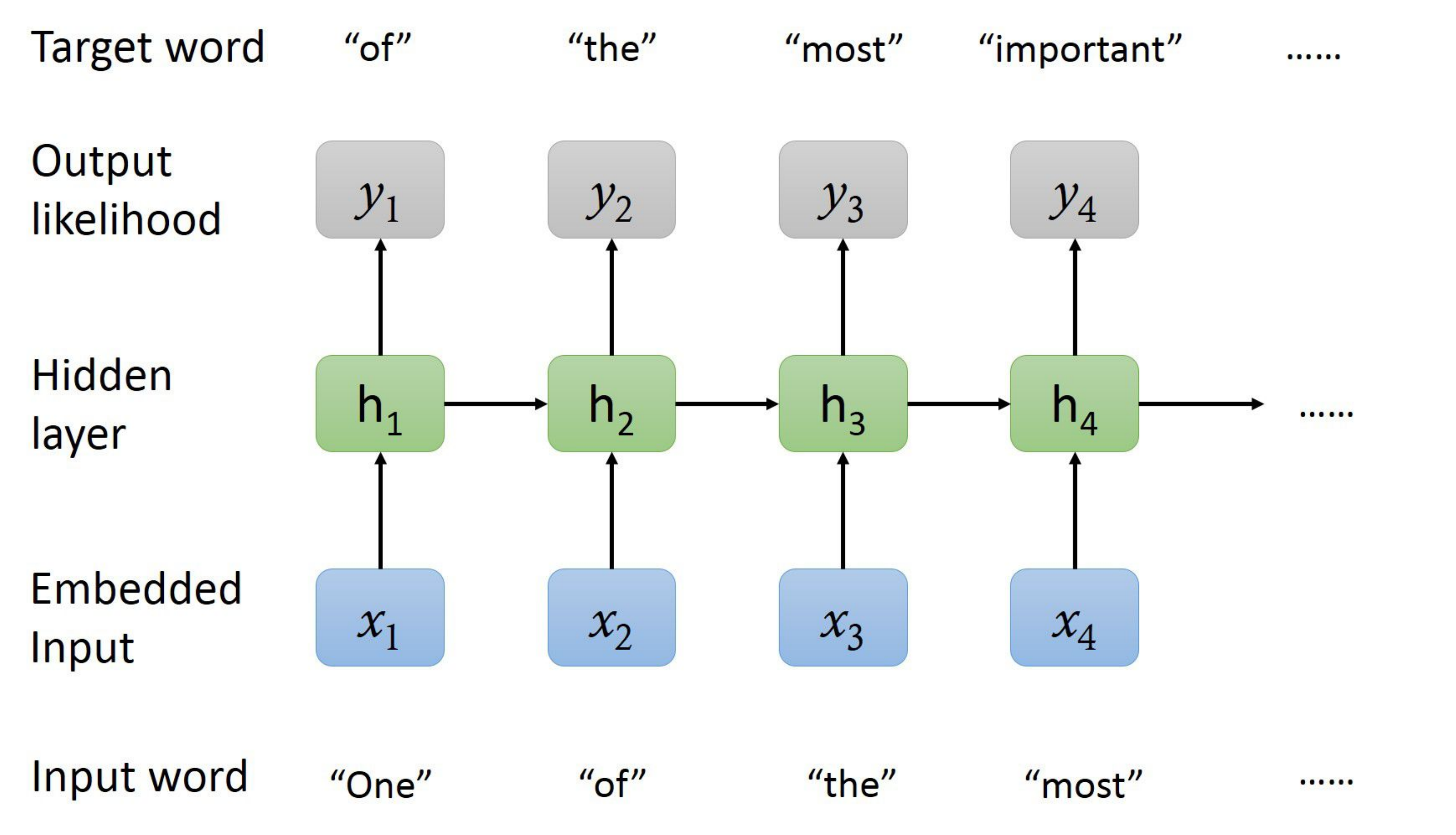}
\end{center}
\caption{Illustration of recurrent neural network language model}
\label{f.rnn}
\vspace*{-3pt}
\end{figure}

Unlike feedforward neural networks, recurrent neural networks (RNNs) make use of sequential information in predicting output \cite{rnn}. 
In another words, RNNs' inference does not solely depend on the current input, but also on information from the previous state.
This is similar to human decision-making, in which a case is judged not only using current information, but also using past experience. 
The RNN language model (RNNLM) is a network model used for predicting a sentence in language and an example is presented in Fig. \ref{f.rnn} \cite{rnnlm}. 
By carrying previous information to the next layer, the model links all input and predicts the next word with the highest possibility of appearing after the current word.

In the RNNLM, perplexity is used instead of accuracy to measure performance as in Eq. \ref{eq.perplexity}.
The perplexity of discrete probability distribution $p$ is defined such that $H(p)$ is the entropy (in bits) of the distribution and $x$ ranges over events \cite{perplexity}.
The model is trained to minimize the cross entropy $H(p)$ to increase performance.
In other words, the smaller the perplexity, the better the RNNLM.

\begin{equation} 
 2^{H(p)} = 2^{-\sum_{x} p(x)  \log_{2} p(x)}
 \label{eq.perplexity}
\end{equation} 

\subsection{Dropout}
\begin{figure}
\begin{center}
\includegraphics[width=8cm]{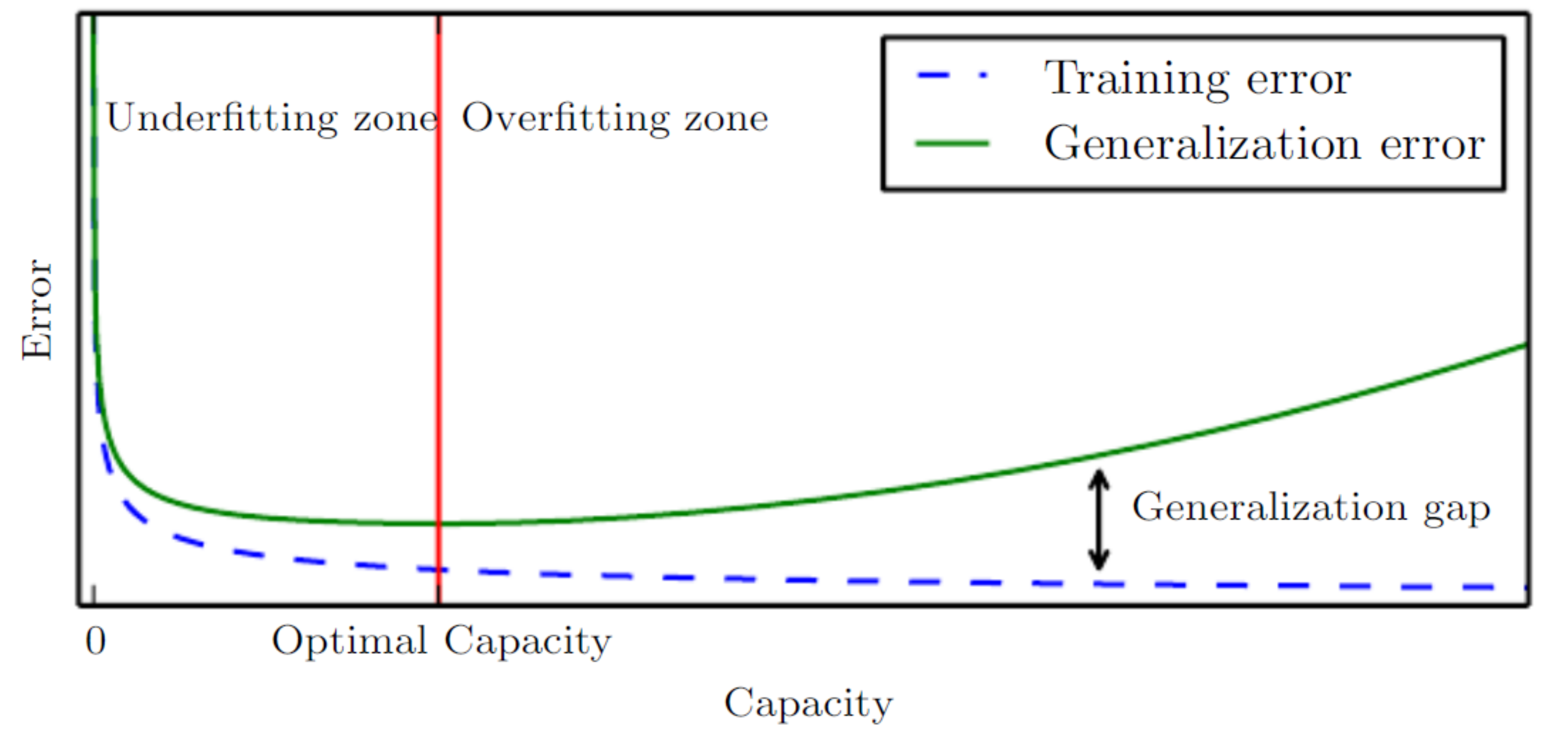}
\end{center}
\caption{Overfitting problem in a neural network}
\label{f.overfitting}
\vspace*{-3pt}
\end{figure}

\begin{figure}
\begin{center}
\includegraphics[width=6cm]{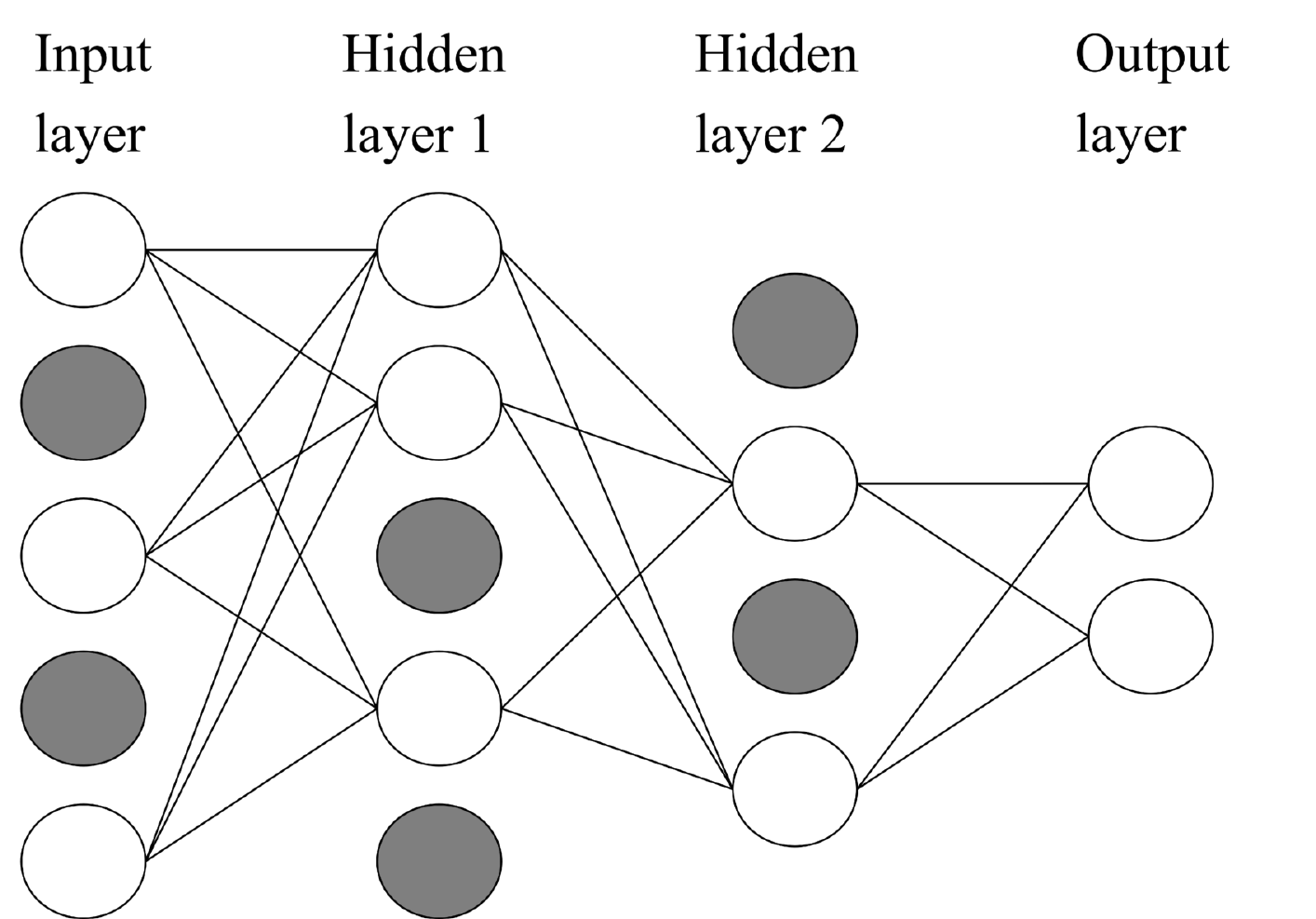}
\end{center}
\caption{Illustration of dropout in neural network}
\label{f.dropout}
\end{figure}

In neural networks, overfitting is a common problem during training. 
As illustrated in Fig. \ref{f.overfitting}, the network is trained to learn and fit the training data.
Consequently, although the training error is minimized, the generalization gap and error are increased over time.
Dropout is a regularization technique used to overcome this problem \cite{dropout}.
This technique randomly drops neurons in the network with a certain ratio to avoid complex co-adaptions between neurons and to learn more robust features as shown in Fig. \ref{f.dropout} \cite{dropout2, dropout3}.
In addition, because training neural networks with dropout deactivates neurons with a certain ratio, this eventually decreases the size of networks, resulting in sparser networks. As a result, the computational cost is decreased \cite{fpgarbm}.
To apply the dropout technique, a dropout mask is attached to each layers of the network.
The dropout mask is a binary vector with a Bernoulli distribution, as in Eq. \ref{eq.mask} and \ref{eq.dropout}.
The neurons are dropped when multiplied by 0, whereas neurons multiplied with 1 will remain as same and are fed forward to next layer.
The dropout masks are changed for every input. 

\begin{equation}
  \emph{mask}_{i}^{(l)} \sim Bernoulli (p) \; 
  \label{eq.mask}
\end{equation}

\begin{equation}
  y_{j}^{(l+1)} = f \lbrace \textbf{\emph{W}} _{j}^{(l+1)} \cdot (\textbf{\emph{mask}}^{(l)} \ast \textbf{\emph{x}}^{(l)}) + b_{j}^{(l+1)} \rbrace \;
  \label{eq.dropout}
\end{equation}

The dropout method has demonstrated effectiveness in solving the overfitting problem for various networks \cite{dropout, dropout3}, and several studies have further widened its application \cite{fastdropout, dropconnect, spectraldropout}.
One exteneded dropout method, Dropconnect, drops the connections (weights) between neurons rather than the neurons themselves \cite{dropconnect}.
Several studies have also implemented dropout in an FPGA \cite{fpgarbm, ssdropout}.
In a study of Su et. al., a restricted boltzmann machine (RBM) with dropout was implemented on an FPGA.
Instead of comparing random numbers to dropout ratio, only $HSd$ random numbers were generated and the dropout ratio was determined by $HSd/HS$, where $HS$ was the size of neuron layer \cite{fpgarbm}.
The random numbers were used as the indices of column in weight matrix, to address the selected weight which stored in external memory \cite{fpgarbm}.
The serial process of RNG and the transfer cost of weight between RNG, external memory and on-chip RAM is concerned.
In a study of Sawaguchi et. al., slightly-slacked dropout was proposed, to alleviate the transfer cost of the method of Su et. al. and accelerate the training \cite{ssdropout}.
A neural central controller was introduced to control the neuron group (NG) and combination (NC) information of slightly-slacked dropout \cite{ssdropout}.
Four subsequent neurons formed a group (NG) with a certain dropout rate, and the approximation of dropout rate was computed across all groups \cite{ssdropout}.
The dropout rate of each NG can only be set to 1, 0.5 or 0, where when the dropout rate was set to 0.5, two neurons will be chosen out of four patterns (fixed) which was controlled using 2-bits (NG Info) \cite{ssdropout}.
The 2-bits NC info were used to assign one out of four combination of the dropout rate of either one, two or three NGs \cite{ssdropout}.
However, these approaches have limitations, such as accuracy degradation, transfer costs, and problems as operating the dropout technique externally between software and hardware \cite{fpgarbm, ssdropout}.
In this paper, we propose an alternative approach that fully enables the application of dropout in hardware with parallel processing in order to address the problem of transfer costs.
In addition, resources are reduced by eliminating RNG.
The proposed method is compared to the conventional dropout method as a baseline.

\section{Proposed Method}
The current dropout approach is primarily a software-based implementation.
To generate a dropout mask, which is a binary vector, a random number generated from an RNG is compared to the dropout ratio which is usually set to 0.5 by a user for a hidden layer \cite{dropout}.
If the random number is larger than the dropout ratio, the value of the dropout mask is set to 1, and if the random number is smaller than the dropout ratio, the value is set to 0 \cite{dropout}.
This process is looped until a binary vector (equal in size to the corresponding layer) is generated.
The dropout mask is generated differently for each layer of the network, and regenerates with changed input.
In software, a random number is easily generated from a program, and executed in a fast clock cycle, thus the looping process does not have a large impact.
However, the looping process significantly increase the burden of execution in an FPGA due to different architecture. 
The serial looping process is not preferred in an FPGA and is ineffective for FPGA implementation, as RNGs and comparators are required, which consume significant resources.
To process in parallel, the number of RNG and comparator will eventually increase, further consuming FPGA 
resources.

\begin{figure}
\begin{center}
\includegraphics[width=8cm]{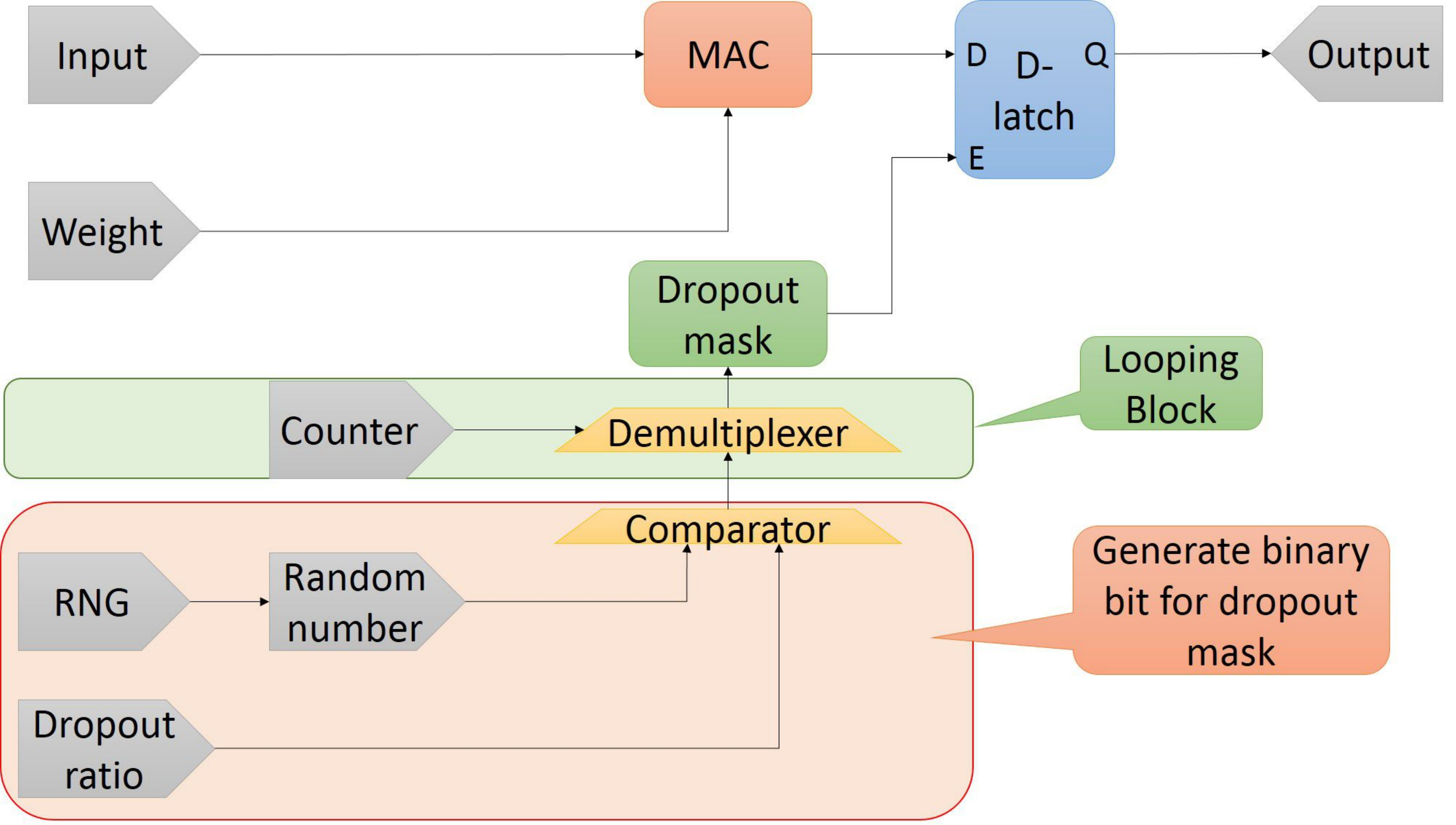}
\end{center}
\caption{Block diagram of general dropout in hardware implementation. (MAC: multiply-accumulate)}
\label{f.block}
\vspace*{-3pt}
\end{figure}

\begin{figure}
\begin{center}
\includegraphics[width=12cm]{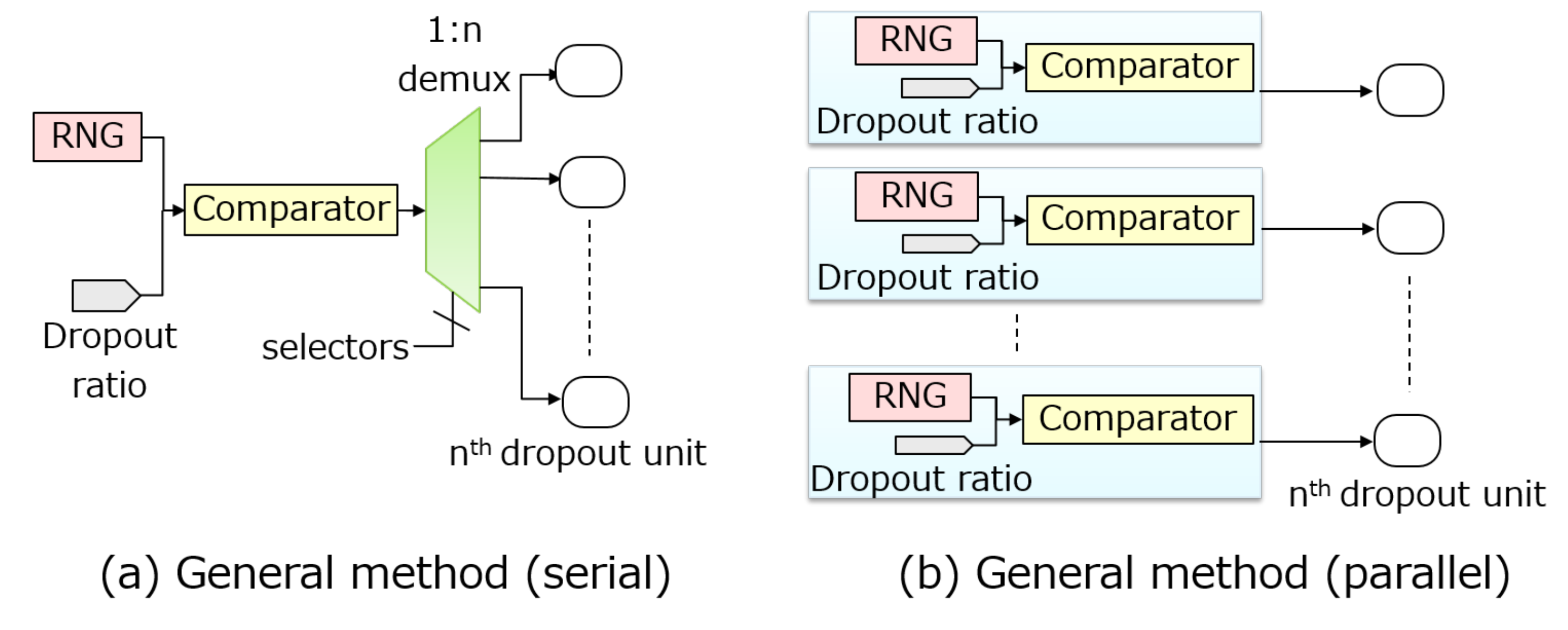}
\end{center}
\caption{Block diagram of comparator block in serial and in parallel}
\label{f.general}
\vspace*{-3pt}
\end{figure}

Figure \ref{f.block} presents a block diagram of general dropout in hardware implementation with serial processing.
The input neurons and weight parameters are summed and multiplied with multiply-accumulate units, while at the same time, the dropout mask is generated by comparing the random number and dropout ratio.
A looping block is required, as only one bit is generated in the comparison.
The dropout mask is connected to the enable of the D-latch, which controls the activation of the neurons. 
The looping block (green) can be eliminated by multiplying the RNG and comparator block and processing the comparison in parallel; however, resource consumption significantly increases, as illustrated in Fig. \ref{f.general}.

Thus, to efficiently apply the dropout method in hardware, we propose an alternative approach.
The proposed method eliminates the use of an RNG and comparator; instead, a predefined mask is used with the addition of a control block and reconfiguration block.
The proposed method not only saves resources by eliminating the use of an RNG and comparator, but also processes in parallel, allowing the regeneration of the dropout mask in a single clock cycle.
In general, dropout drops a neuron with true randomness, whereas the proposed method drops neurons with pseudorandomness, which is considered sufficient for most applications even though the randomness has a pattern and is predictable \cite{rng}.
It has been hypothesized that true randomness in dropout is not essential, and that dropout can perform well even with pseudorandomness \cite{yeoh}.

\begin{figure}
\begin{center}
\includegraphics[width=10cm]{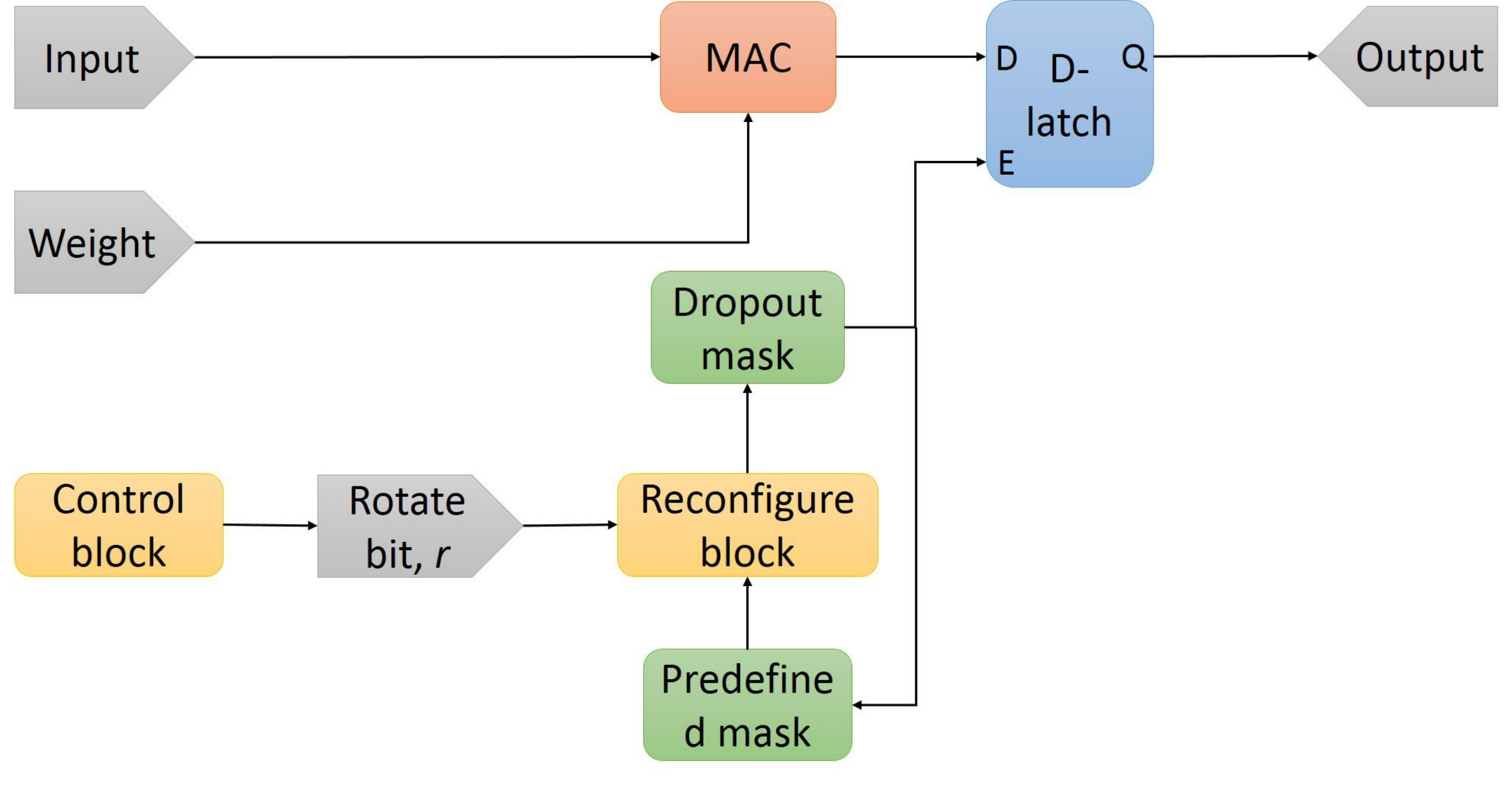}
\end{center}
\caption{Block diagram of proposed dropout method}
\label{f.proposedblock}
\vspace*{-3pt}
\end{figure}

\begin{figure}
\begin{center}
\includegraphics[width=5cm]{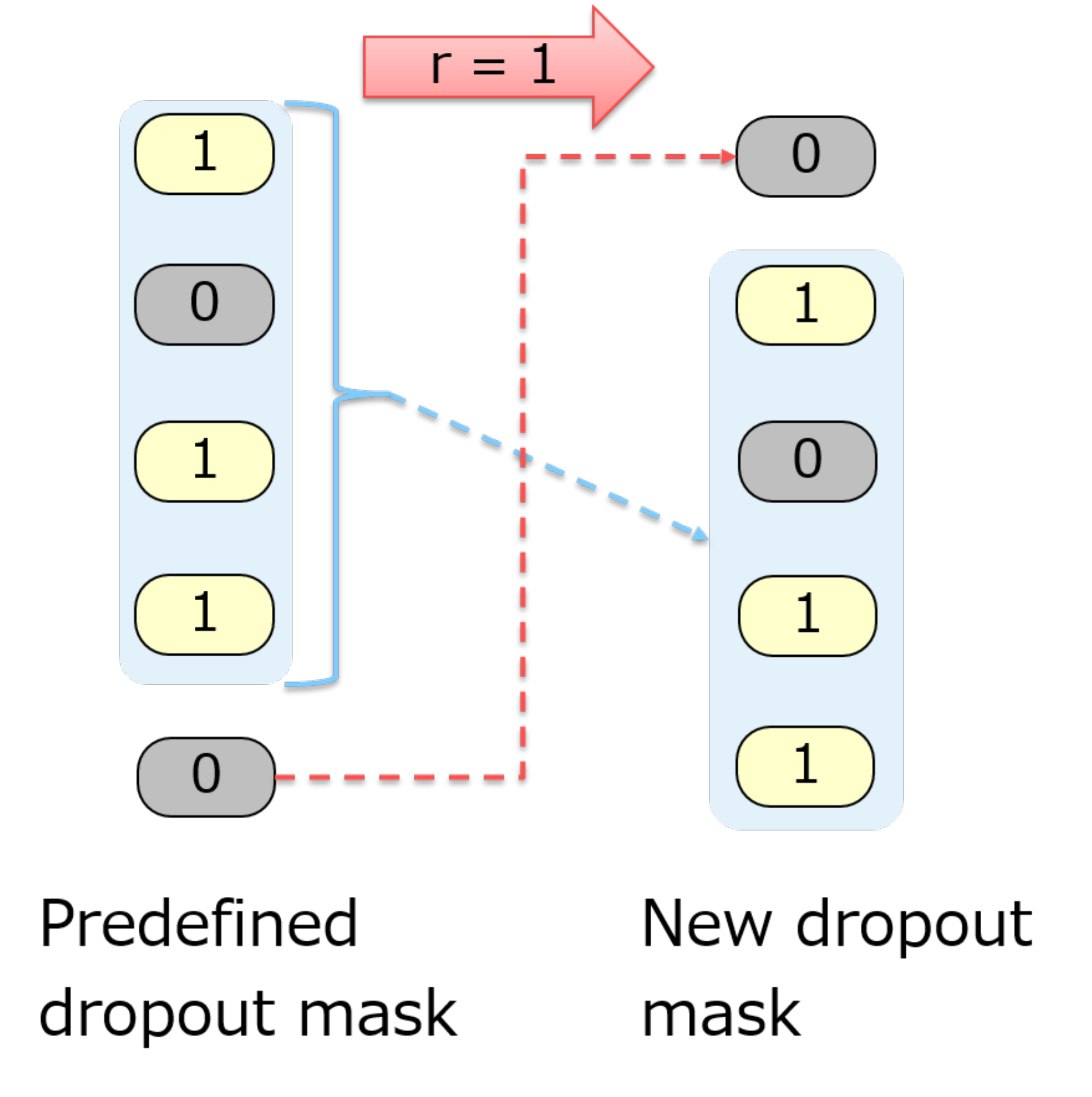}
\end{center}
\caption{Configuration block of proposed dropout method}
\label{f.proposed}
\vspace*{-3pt}
\end{figure}

\begin{figure}
\begin{center}
\includegraphics[width=10cm]{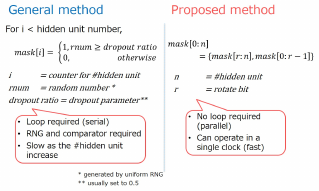}

\caption{Comparison between general dropout algorithm and proposed algorithm}
\label{f.algorithm}
\vspace*{-3pt}
\end{center}
\end{figure}

A predefined mask is generated and saved in memory for initialization purposes.
For each generation of the dropout mask, a predefined mask is loaded to the reconfiguration block to reconstruct the mask, changing the sequence of the mask as a new dropout mask.
In the reconfiguration block, a simple, parallel operation is executed. 
For simplicity, the experiments in this study applied rotation in the reconfiguration block, whereas in the control block, the parameter of the rotate bit, $r$, was used to control the bit rotation in the configuration block.
By only rotating bits of the mask, the distribution is maintained and the operation remains simple.
In this paper, the effect of parameter $r$ was studied by setting it to be constant, a sequence, and random.
Instead of an RNG and comparator, the reconfiguration block and control block were used to consume fewer resources with high processing speed.
Figure \ref{f.proposedblock} presents a block diagram of the proposed method, while Fig. \ref{f.proposed} presents a simple illustration of the rotation in the configuration block.
A new dropout mask can be generated by rotating the bit of the dropout mask in parallel.
The figure illustrates that no looping process is required in the proposed method.
In Fig. \ref{f.algorithm}, algorithms for general dropout and the proposed method are compared.
As demonstrated in the figure, the serial looping for general dropout is effective in hardware, as the clock speed is slow, and the clock cycle required is proportional to the number of neurons. 
In contrast, the proposed method is simple and operates in parallel in a single clock cycle with fewer resources. 

\section{Experimental Results}
We conducted several experiments to verify the effectiveness of the proposed method.
The experiments had two main approaches: software program verification and hardware resources analysis.
In software program verification, the algorithm of the proposed method was examined and compared with that of general dropout. 
Various types of neural networks with different datasets were used to test the effect and robustness of the proposed method.
In hardware resources analysis, the resources consumed in generating the dropout masks were compared, to observe the amount of resources that could be preserved.  

\subsection{Software program verification}
Four different types of network architectures with four different datasets were tested, as follows:
\begin{enumerate}
\item MLP - MNIST
\item LeNet - CIFAR10
\item GoogLeNet - @home dataset
\item RNNLM - Penn Treebank (PTB)
\end{enumerate}
The MLP network was trained using the C language, whereas the other networks were trained and tested using the Chainer platform \cite{chainer}.
Each network was trained for five trial to ensure its robustness, and the average results were obtained and plotted as a figure.
The performance of the conventional method and proposed method was compared in terms of classification accuracy.

\subsubsection{\textbf{MLP - MNIST}}
The first network implemented was a four-layer MLP, which was 784-500-200-10 as illustrated in Fig. \ref{f.mlp}.
The MLP was trained with the MNIST dataset, which is a dataset of handwritten number images \cite{cnn}.
Each image in MNIST was a grayscale image with a size of 28 x 28 pixels (784 pixels).
It contained 10 classes (from 0 to 9), and a total of 60,000 training data images and 10,000 test data images.
To train the MLP, mini-batch stochastic gradient descent (SGD) with a batch size of 100 was used as the optimizer.
Softmax regression was used as the cost function.

\begin{figure}
\centering
\includegraphics[width=10cm]{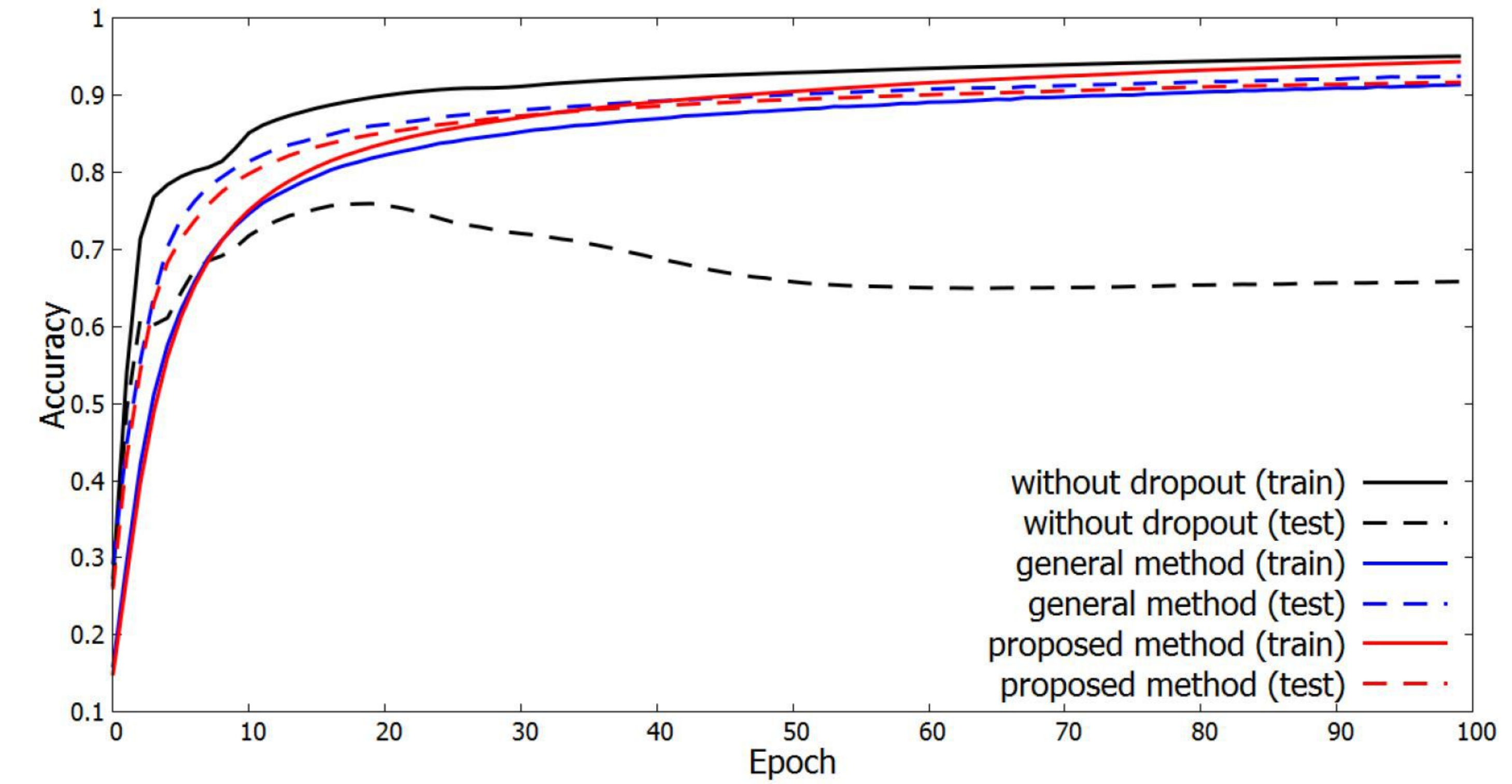}
\caption{Results of multi-layer perceptron (MLP) trained with MNIST dataset.}
\label{f.MNISTresult}
\end{figure}

The training and test accuracy of the three approaches is presented in Fig. \ref{f.MNISTresult}. 
The overfitting phenomenon can be clearly observed for the MLP that did not apply dropout (black lines).
The training accuracy (solid- line) was close to 95\%, while the test accuracy (dotted- line) was saturated at approximately 65\%, revealing a large gap between the two.
The MLP failed to predict new test data due to overtraining and overfitting to the training data.
In contrast, the general dropout method (blue lines) was able to avoid the overfitting problem and achieve accuracy of over 90\% during the training and inference phase.
The proposed method (red lines) performed well and achieved a similar effect to the general dropout method.

\subsubsection{\textbf{LeNet}}
\begin{figure}
\centering
\includegraphics[width=10cm]{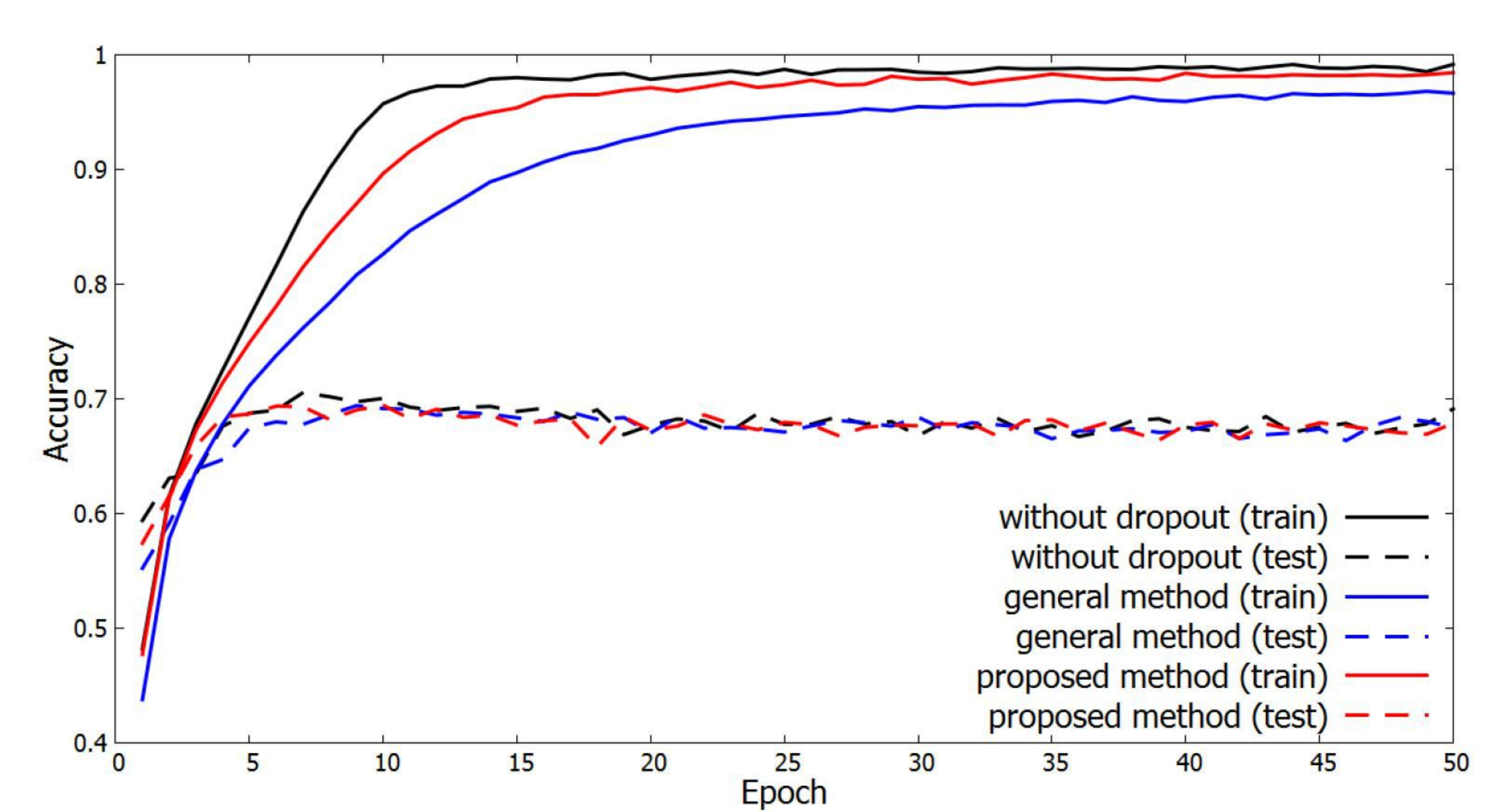}
\caption{Results of LeNet trained with CIFAR10 dataset.}
\label{f.CIFAR10result}
\end{figure}

Here, the proposed method was evaluated with LeNet, a CNN with the architecture illustrated in Fig. \ref{f.cnn}.
Adam optimization was used, and dropout was only applied in the fully-connected layer to avoid unnecessary loss.
Spatially shared- weights in a CNN have relatively few parameters and generally do not cause overfitting.
The dataset used in this experiment was CIFAR10, a dataset containing 10 classes of objects, including an airplane, automobile, and bird \cite{cifar10}.
CIFAR10 contained 50,000 images as training data and 10,000 images as test data.
Each image size was 32 x 32 pixels and consisted of RGB, three color channels.

Figure \ref{f.CIFAR10result} demonstrates that all of the approaches had similar results; the training accuracy was over 90\%, while the test accuracy was saturated at approximately 70\%.
Although dropout was applied for LeNet, overfitting still occurred.
This may be because LeNet is a shallow CNN and may not perform well with the CIFAR10 dataset, which comprises color image data that are relatively difficult to classify.
However, the main objective of this paper is not to examine the effect of dropout, but rather to compare the proposed method with conventional approaches. 
Therefore, although dropout did not demonstrate favorable results in this experiment, the proposed method was observed to have identical results to general dropout.

\subsubsection{\textbf{GoogLeNet}}
This experiment involved a deeper CNN called GoogLeNet which consists of 22 network layers and took first place for object classification in the ImageNet Large- Scale Visual Recognition Challenge (ILVRC) in 2014 with excellent recognition performance \cite{ilsvrc1, ilsvrc2}.
The @home dataset was created for the RoboCup Japan Open 2016 @Home competition \cite{athome, dataset}.
The CNN was trained with the dataset and implemented into a home service robot to allow the robot to recognize objects and perform service tasks. 
The @home dataset was a 15 object class dataset, in which each class contained of 2,000 training data images and 700 test data images, as illustrated in Table \ref{t1}.
Each image was resized to 224 x 224 pixels with three color channels.
To match the dataset, the output layer of GoogLeNet was modified to 15 neurons instead of 1,000 neurons.

\begin{table}
\centering
\caption{Object classes in @home dataset}
\label{t1}
\begin{tabular}{llll}
\hline \noalign{\smallskip} 
Item Categories  & Examples \\
\noalign{\smallskip}
\hline\noalign{\smallskip}
PET bottle  & \includegraphics[width=0.7cm, height=0.9cm]{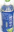} & \includegraphics[width=0.7cm, height=0.9cm]{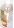} & \includegraphics[width=0.7cm, height=0.9cm]{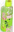} \\
 & Iced-tea & Cafe-Au Lait & Green tea \\
\noalign{\smallskip}
\hline\noalign{\smallskip}
Snack & \includegraphics[width=0.7cm, height=0.7cm]{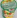} & \includegraphics[width=0.7cm, height=0.7cm]{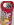} & \includegraphics[width=0.7cm, height=0.7cm]{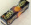} \\
 & Potato stick & Potato chips & Chocolate cookies \\
\noalign{\smallskip}
\hline\noalign{\smallskip}
Fruit juice & \includegraphics[width=0.7cm, height=0.7cm]{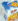} & \includegraphics[width=0.7cm, height=0.7cm]{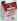} \\
 & Orange juice & Strawberry juice \\
\noalign{\smallskip}
\hline\noalign{\smallskip}
Fruit & \includegraphics[width=0.7cm, height=0.7cm]{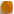} & \includegraphics[width=0.7cm, height=0.7cm]{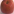} \\
 & Orange & Apple \\
\noalign{\smallskip}
\hline\noalign{\smallskip}
Instant soup & \includegraphics[width=0.7cm, height=0.7cm]{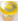} & \includegraphics[width=0.7cm, height=0.7cm]{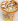} \\
 & Egg soup & Potato soup \\
\noalign{\smallskip}
\hline\noalign{\smallskip}
Container & \includegraphics[width=0.7cm, height=0.7cm]{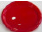} & \includegraphics[width=0.7cm, height=0.7cm]{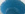} & \includegraphics[width=0.7cm, height=0.7cm]{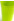} \\
 & Tray & Bowl & Cup \\
\noalign{\smallskip}
\hline\noalign{\smallskip}
\end{tabular}
\end{table}

\begin{figure}
\centering
\includegraphics[width=10cm]{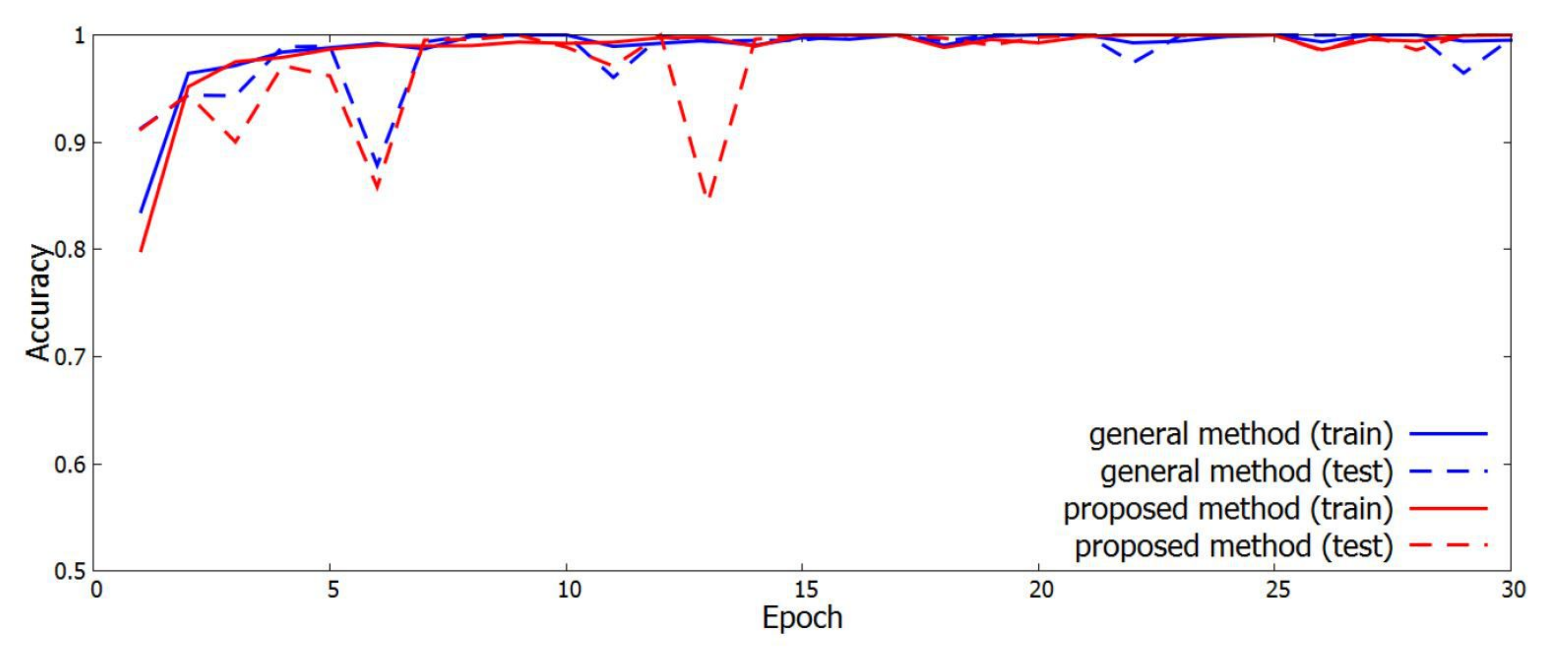}
\caption{Results of GoogLeNet trained with @home dataset.}
\label{f.athomeresult}
\end{figure}

The main concern of this study is to compare and evaluate the performance between general dropout and the proposed method, thus the result of GoogLeNet without dropout is not demonstrated.
The performance of general dropout and the proposed method is compared in Fig. \ref{f.athomeresult}.
Both approaches achieved stable and recognition accuracy close to 100\% after 15 epochs.
The high accuracy in both the training and inference phases indicates that the network was trained well and no overfitting occurred.
The deep CNN was highly effective for the @home dataset, and the proposed method achieved the same effect as the general method.

\begin{figure}
\begin{center}
\includegraphics[width=10cm]{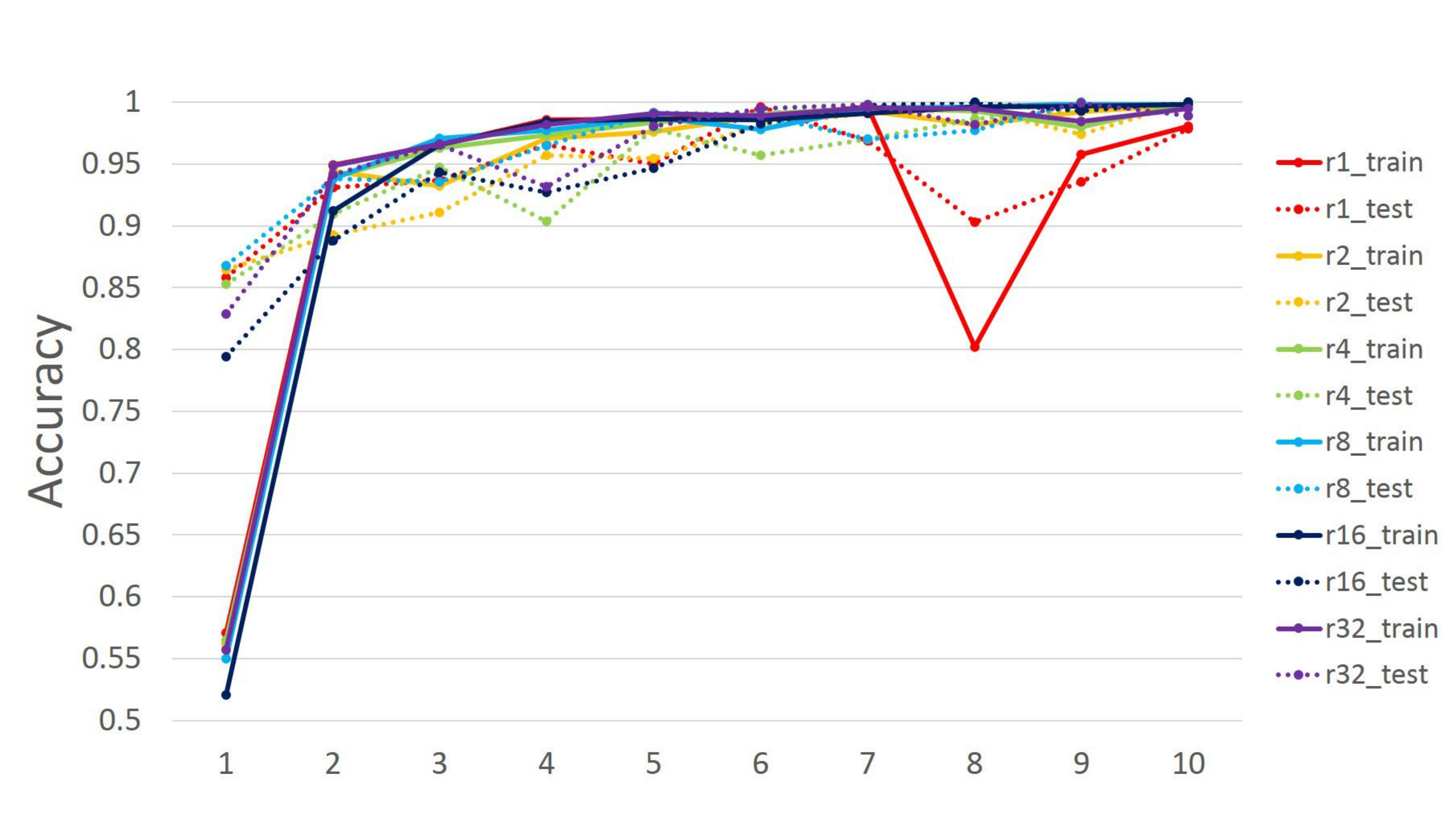}
\end{center}
\caption{Effect of proposed method with varied parameter $r$}
\label{f.rotatebit}
\vspace*{-3pt}
\end{figure}

In the proposed method, the dropout mask length is equal to the number of neurons in each respective layer.
Parameter $r$ is adjustable and was set to a sequential constant number in the previous experiments.
The effect of changing $r$ is another concern; thus, with the same environment settings as GoogLeNet experiment, $r$ was changed to a constant number, 1, 2, 4, 8, 16, and 32, to observe the produced effect.
Figure \ref{f.rotatebit} indicates that the accuracy was relatively unstable when $r$ was set to 1; however, it did not significantly change when $r$ was increased.
Thus, $r$ can be defined as a constant or varying function depending on the available resources.

\subsubsection{\textbf{RNN}}
\begin{figure}
\begin{center}
\includegraphics[width=8cm]{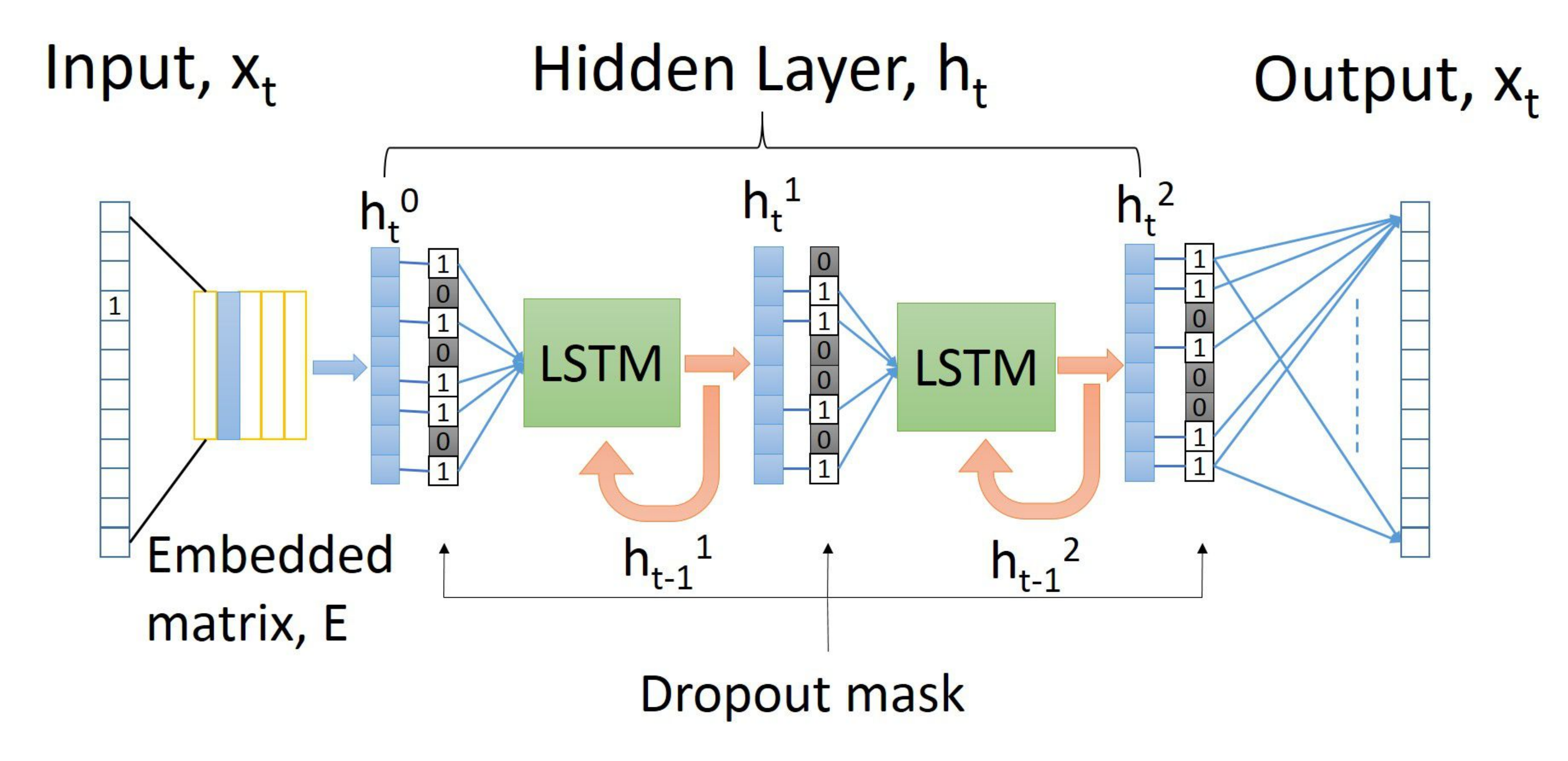}
\end{center}
\caption{Structure of recurrent neural network language model (LSTM: long-short-term memory)}
\label{f.rnnlm}
\vspace*{-3pt}
\end{figure}

In addition to examining feedforward neural networks, the experiments were extended to verify the effectiveness of an RNN.
RNNLM was implemented and trained with the Penn Treebank (PTB) dataset, which contains 10,000 vocabulary words in sequence and is commonly used in natural language processing \cite{ptb}.
Words corresponding to their ID were input as a one-hot vector that was embedded into an embedded matrix before inputting to RNN.
The RNNLM used in this experiment consisted of two hidden layers with long-short-term memory (LSTM), and the dropout layer was applied before the LSTM layer.
The structure of the RNNLM was composed by five layers (10,000-650-650-650-10,000), as illustrated in Fig. \ref{f.rnnlm}.

\begin{figure}
\begin{center}
\includegraphics[width=12cm]{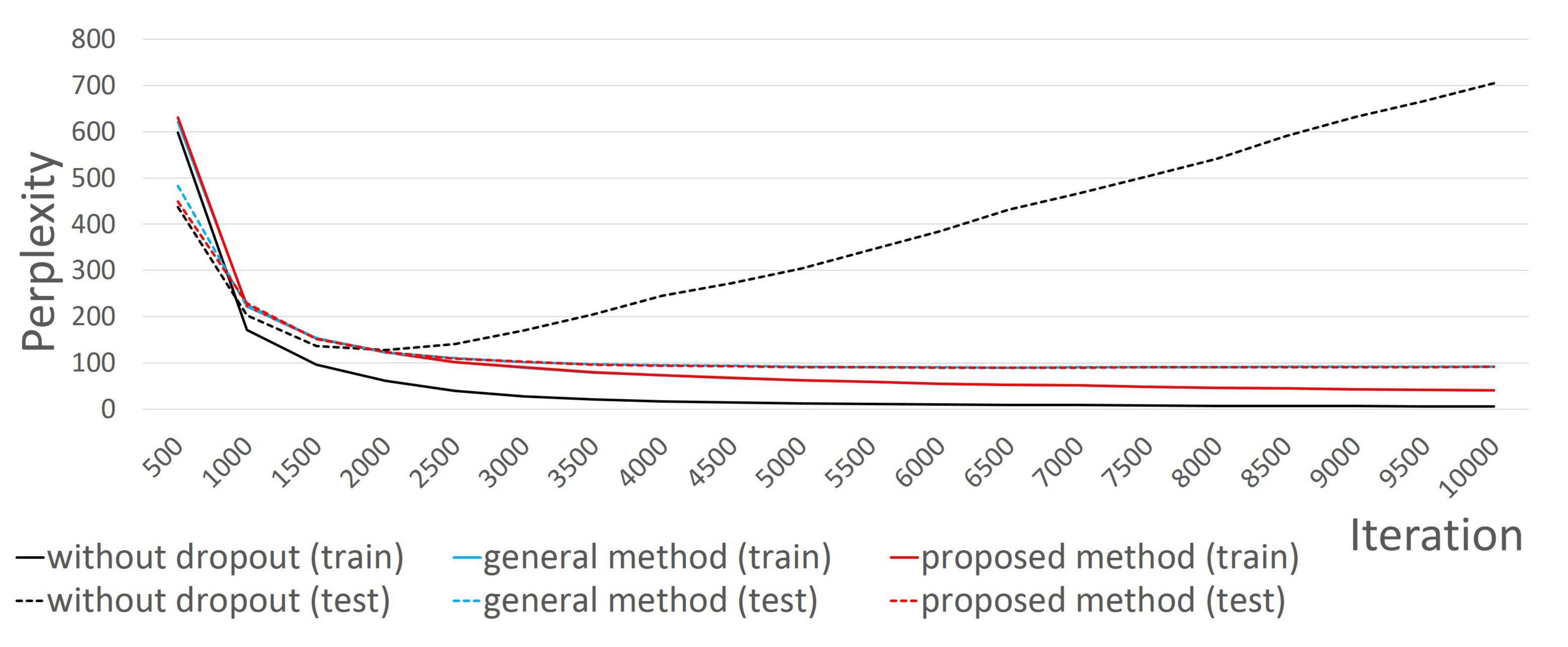}
\end{center}
\caption{Comparison of recurrent neural network language model}
\label{f.rnnresult}
\vspace*{-3pt}
\end{figure}

When dropout was not applied, overfitting occurred, as indicated in Fig \ref{f.rnnresult}.
The test perplexity increased while the training perplexity remained low.
However, when dropout was applied to the RNNLM, the gap between training perplexity and test perplexity was reduced.
Table \ref{t.rnn} also illustrates that the test perplexity reached over 750, which was in contrast to approaches using dropout, whose perplexity values were only approximately 89.
The results of the proposed method were thus identical to those of the general dropout method.

\begin{table}
\begin{center}
\begin{threeparttable}
\caption{Comparison of test perplexity between different methods}
\label{t.rnn}
\begin{tabular}{|c|c|}
\hline
\centering \bf{Approaches} & \bf{Test perplexity} \\
\hline
Without dropout & 757.67 \\
General dropout & 88.99 \\
Proposed dropout & 89.96 \\
\hline
\end{tabular}
\end{threeparttable}
\end{center}
\end{table}

\subsection{Hardware resources analysis}
After verifying the effectiveness of the proposed method, we examined resources saving.
Hardware resource analysis was performed to compare the amount of resources required using the general method and proposed method.
An experiment was conducted by implementing the dropout mask only.
We generated 8- and 64- bit dropout masks using the general method in series, general method in parallel, and the proposed method.
In actual application, the size of the dropout mask is proportional to neurons in the layers, and multiple dropout masks are required for different layers of neural network.
Thus, it is expected that more resources would be saved for actual application than are presented in Table \ref{t.compare1} and \ref{t.compare2}.

In Table \ref{t.compare1}, the resources consumed by an 8-bit dropout mask generated by different approaches are tabulated through hardware synthesis.
The most resources were consumed with general method with parallel processing due to the duplication of RNGs and comparator blocks.
For the general method with serial processing, fewer resources were required; however, a longer time and clock cycle were required.
The proposed method had the advantages of the other two approaches, as it consumed the fewest resources and required only a single clock cycle to generate the mask.
Similarly, a comparison was performed between the three approaches in generating a 64-bit dropout mask, as presented in Table \ref{t.compare2}.
As indicated in the table, a large number of resources was required with the serial approach, which was further increased with the parallel approach.
Fewer resources were consumed with the proposed method, and only a single clock cycle was required.
Therefore, the proposed method is effective in hardware implementation and is characterized by low resource consumption and a fast processing speed.

\begin{table}
\begin{center}
\centering
\caption{Comparison of field programmable gate array resources consumed for 8-bit dropout masks}
\label{t.compare1}
\begin{tabular}{|p{30mm}|p{15mm}|p{15mm}|p{15mm}|c}
\hline \centering
\bf{Logic utilization} & \bf{General dropout (serial)} & \bf{General dropout (parallel)} & \bf{Proposed method} \\
\hline
\bf{Number of slice} & 32 & 7 & 8 \\
\bf{registers} & & & \\
\hline
\bf{Number of slice LUTs} & 44 & 80 & 7 \\
\hline
\bf{Number of fully used LUT-FF pairs} & 27 & 72 & 0 \\
\hline
\bf{Clock cycle required to generate a mask} & 8 & 1 & 1 \\
\hline
\bf{RNG required} & 1 & 8 & N/A \\
\hline
\end{tabular}
\end{center}
\end{table}

\begin{table}
\begin{center}
\centering
\caption{Comparison of field programmable gate array resources consumed for 64-bits dropout masks}
\label{t.compare2}
\begin{tabular}{|p{30mm}|p{15mm}|p{15mm}|p{15mm}|c}
\hline \centering
\bf{Logic utilization}  & \bf{General dropout (serial)} & \bf{General dropout (parallel)} & \bf{Proposed method} \\
\hline
\bf{Number of slice} & 149 & 588 & 70 \\
\bf{registers} & & & \\
\hline
\bf{Number of slice LUTs} & 190 & 640 & 64 \\
\hline
\bf{Number of fully used LUT-FF pairs} & 141 & 576 & 64 \\
\hline
\bf{Clock cycle required to generate a mask} & 64 & 1 & 1 \\
\hline
\bf{RNG required} & 1 & 64 & N/A \\
\hline
\end{tabular}
\end{center}
\end{table}

\subsection{Discussion}
We made the comparison between the general dropout and proposed dropout in aspect of processing, time, randomness, generation of dropout mask, resources required and its suitability device, as tabulated in Table \ref{t.summary}.
The general dropout is a serial processing algorithm where the proposed dropout designed for parallel processing.
Thus, the proposed dropout is faster than general dropout and independent to the number of neurons (the size of dropout mask).
However, the proposed method has lower randomness as it generates dropout masks based on a predefined mask initially.
In aspect of resource, as the general dropout requires RNGs and comparators, thus the resource consumed is high;
Where the proposed dropout requires only very small amount of resource.
For software application such as CPU, the general dropout is more suitable, whereas for hardware devices such as FPGA, the proposed dropout has advantages to it and is efficient for the implementation.

\begin{table}
\begin{center}
\begin{threeparttable}
\caption{Summary of general dropout and proposed dropout}
\label{t.summary}
\begin{tabular}{|p{30mm}|p{30mm}|p{30mm}|c}
\hline
\centering  & \bf{General dropout} & \bf{Proposed dropout} \\
\hline
Processing & Serial looping & Parallel \\
\hline
Time & Slow (dependent on \# of neurons) & Fast (independent of \# of neurons) \\
\hline
Randomness & High & Normal \\
\hline
Dropout mask & Regenerated for each layer & Predefined \\
\hline
Resources Required & Very high (for parallel) & Low \\
 & high (for serial) & \\
\hline
Suitable application & Software & Hardware \\
\hline
\end{tabular}
\end{threeparttable}
\end{center}
\end{table}

\section{Conclusions}
We demonstrated that our proposed dropout method has the same performance as the general dropout method and it is characterized by parallel processing and greater resource savings.
The proposed method is effective in hardware implementation with its parallel processing characteristic and 
Using the proposed method, the dropout technique can be applied with fewer resources without reducing the performance of the neural network.
The number of resources can be further reduced as the number and size of the dropout masks increase.
The saved resources can be utilized in other algorithms in neural networks, overcoming the memory constraint in FPGAs.

\section*{Acknowledgment}
This research was supported by JSPS KAKENHI Grant Numbers 17H01798 and 17K20010.



\bibliographystyle{elsarticle-num}

\bibliography{ref}





\end{document}